\documentclass{article}
\usepackage{acra}
\usepackage{amsmath}     
\usepackage{amssymb}
\usepackage{xcolor}   
\usepackage{float}   
\usepackage{algpseudocode}
\usepackage{algcompatible}
\usepackage{float} 
\usepackage{algorithmicx}
\usepackage[ruled,vlined,linesnumbered]{algorithm2e}
\usepackage{booktabs}
\usepackage{subcaption}
\usepackage[hidelinks]{hyperref} 

\title{ Optimizing Robotic Placement via Grasp-Dependent Feasibility Prediction}
\author{\normalfont
  Tianyuan Liu, Richard Dazeley, Benjamin Champion, Akan Cosgun
  \thanks{All authors are with Deakin University, Australia.
  Email: \{tianyuan.liu, richard.dazeley, benjamin.champion, akan.cosgun\}@deakin.edu.au}
}

\begin{document}

\maketitle
\begin{abstract}
In this paper a study of whether inexpensive, physics-free supervision can reliably prioritize grasp–place candidates for budget-aware pick-and-place is presented. From an object's initial pose, target pose, and a candidate grasp, two path-aware geometric labels were generated: path-wise inverse kinematics (IK) feasibility across a fixed approach–grasp–lift waypoint template, and a transit collision flag from mesh sweeps along the same template. A compact dual-output MLP learns these signals from pose encodings, and at test time its scores rank precomputed candidates for a rank-then-plan policy under the same IK gate and planner as the baseline. Although learned from cheap labels only, the scores transfer to physics-enabled executed trajectories: at a fixed planning budget the policy finds successful paths sooner with fewer planner calls while keeping final success on par or better. This work targets a single rigid cuboid with side-face grasps and a fixed waypoint template, and we outline extensions to varied objects and richer waypoint schemes.
\end{abstract}

\section{Introduction}
Placing an object reliably after grasping is as critical as grasping itself. A successful placement requires choosing a grasp that achieves the target pose, remains reachable at the endpoint, and admits a collision-free path under time and planning limits. Simply separating “pick” from “place” often fails—a grasp that is easy to acquire may be impossible to place without collisions or joint-limit violations.

Much prior work \cite{pointcloud_placement_1,second_reference}treats placement as a minor adjustment to a chosen grasp or assumes stability once the grasp is fixed, which breaks down in constrained scenes or specific grasp–pose pairings. Meanwhile, collecting execution-grounded labels at scale is costly: full-physics and planner-in-loop evaluations are slow and brittle, yielding small datasets and limited generalization. This raises a key question: can we learn a simple score from cheap labels that still predicts real planning and execution outcomes?

The proposed answer is that inexpensive, physics-free supervision can reliably guide executed pick-and-place within a controlled, single-object setting. From an object’s start/target poses and a candidate grasp, two feasibility signals—endpoint reachability and transit collision risk—are learned to rank candidates before planning. In executed simulation, higher-ranked choices succeed more often under the same planning budget. The key contributions are:

\begin{itemize}
  \item \textbf{Scalable, physics-free supervision:} Pick–place evaluation is cast as two fast geometric feasibility signals—path-wise IK over the place sweep and nominal transit collision risk—computed at scale.
   \item \textbf{Compact predictor \& simple policy:} A lightweight model learns these signals and supports a rank-and-plan policy with a standard IK gate at the place grasp.
  \item \textbf{Executed-sim transfer validation:} An external, physics-enabled evaluation that tests whether scores trained on physics-free labels predict executed outcomes.
  \item \textbf{Budget-matched baseline comparison:} Under identical grasp pools, gating, and planner settings, our policy achieves a higher success rate and requires fewer expected planner calls than a rank-agnostic baseline.

\end{itemize}

\section{Related Works}
\noindent In this section, model-based placement in structured settings is firstly contrasted with learning-based approaches that infer from data. Unlike pose-prediction work, the target pose is assumed given, and the question is whether a chosen grasp can feasibly achieve it. Next, joint pick–and–place methods that couple grasp selection with placement are reviewed, highlighting how grasp choice influences reachability and collision risk, and clarifying how this approach differs from prior work.

\subsection{Learning-Based Robotic Object  Placement}
 Object placement can generally be solved with two types of approaches: model-based approach and learning-based approach. At a high level, \emph{model-based} placement reasons from known geometry and physics (e.g., stability analysis and kinematics) to analytically select feasible poses, whereas \emph{learning-based} placement infers predictors from data so that decisions can still be made under partial, noisy, or incomplete object information.

Model-based approaches are used when object models and environments are well specified: they explicitly encode stability and reachability and can give precise placements in structured scenes \cite{harada2014validating,convexhull}. They are also common inside rearrangement pipelines, where the emphasis is on deciding where to put items while respecting robot kinematics and avoiding extra motions in confined workspaces \cite{cheong2020relocate}. The trade-off is that performance hinges on accurate priors (CAD, mass, friction) and clean visibility. Generalization to novel shapes or partial views is limited by design \cite{model_based_limitation}.

Learning-based methods address those gaps by training predictors that map sensor inputs to stable placements without requiring complete models. Recent work on unseen object placement learns to detect stable supporting planes directly from partial point clouds, at scale, and demonstrates transfer to real-world objects without per-object tuning \cite{noh2024learning}. Complementary to vision, tactile-centric approaches use in-gripper sensing to infer a corrective rotation that aligns the object with a planar surface before release, improving robustness when vision is occluded during handover or in clutter \cite{lach2023placing}. 

Compared to prior learning-based placement that primarily generates a desirable final pose for the object \cite{noh2024learning,lach2023placing}, we ask a different question: given the object’s start pose, target pose, and a specific grasp, will the pick–and-place execute under kinematic and collision constraints? We learn to score each grasp–place candidate by endpoint reachability at the place pose and nominal transit collision risk, then rank before planning—improving efficiency without changing the planner.

\subsection{Joint Grasping and Placement}
Pick-and–place success depends on compatibility between the grasp and the downstream placement trajectory: a grasp that lifts reliably can still fail at place time due to reachability or collisions along the transit path. Recent work therefore optimizes grasps and placements jointly rather than in sequence. For example, \cite{second_similar} cast joint pick-and–place as a constrained optimization over grasp configuration and placement pose using partial-view point clouds. Their formulation couples a learned grasp-success term with geometric costs and signed-distance collision checks. They report higher place success in clutter than sequential "pick-then-place" pipelines.
\cite{gualtieri2021robotic} study the same coupling from a different angle: a regrasp planner that explicitly reasons about perceptual uncertainty (instance segmentation + shape completion) and selects grasp–place sequences by maximizing predicted execution success, success prediction and uncertainty-aware costs outperform uncertainty-unaware heuristics in both simulation and real robots .

Prior work relevant to our problem includes \cite{cheng2021learning}, who treated grasp and place as a coupled problem and learn to prioritize compatible grasp–place choices under task objectives (e.g., packing/stacking) using point-cloud perception and learned scoring for feasibility. In parallel, \cite{zhao2025anyplace} focuses on predicting diverse \emph{placement poses} from vision (with a VLM proposing candidate regions), grasps are then found by a separate grasp module before motion planning, so grasp–place coupling is handled implicitly by planner filtering rather than in the scoring model itself .

\noindent The approach differs in two respects: (1) precomputed grasps are scored for a specified target pose via learned reachability and transit-safety signals, checking grasp–transit compatibility before planning; and (2) transfer is validated in physics-enabled execution, testing whether geometry-only supervision predicts outcomes under dynamics. Thus, while prior work either optimizes coupled grasp–place costs or predicts placements and relies on the planner to filter, this work instead learns a lightweight geometric compatibility ranker—trained without physics—for budget-aware execution.

\section{Problem Statement}
The focus of the work is learning to predict whether a nominal pick-and–place template—instantiated by a candidate grasp that will be executed by a 7-DoF robot and the object's initial and target poses. The setting is a tabletop scene with a single known cuboid. For each candidate, the inputs are:
\begin{itemize}
    \item Object’s initial world pose,
    \item A grasp pose expressed in the object’s initial frame
    \item Object’s target world pose
\end{itemize}

The goal is to predict whether executing a pick–place template will
\begin{enumerate}
    \item be inverse-kinematics (IK) feasible, and
    \item remain free of collisions.
\end{enumerate}
The model does not generate pick or place poses, it only scores the feasibility of given candidates so they can be ranked.

\subsection{Notations}
In this work, a rigid transform (rotation + translation) that maps coordinates from frame $B$ to frame $A$ is represented as:
\begin{equation*}
    ^{A}T_{B}
\end{equation*}

And there are four frames used in total: \(W\) (world frame), \(O_i\) (initial object frame), \(O_f\) (final object frame), and \(G\) (grasp/tool frame).
The inputs for one candidate are exactly
\[
^W\!T_{O_i},\quad ^{O_i}T_G,\quad ^W\!T_{O_f},
\]
where \(^W\!T_{O_i}\) and \(^W\!T_{O_f}\) are the object poses in world at pick phase and place phase,
and \(^{O_i}T_G\) is the grasp pose expressed in the initial object frame. Object’s initial and target poses vary in both position and orientation.  It is assumed that no slipping/rotation after the robot grasps the object, so the \(^{O_i}T_G\) will remain constant during the whole trajectory path. For convenience the grasps in world at each phase can be represented as:
\[
^{W}\!T_{G_i} \;=\; ^W\!T_{O_i}\,^{O_i}T_G \;=\
^{W}\!T_{G_f}
\]

\subsection{Deterministic motion primitives}

Given \((^W\!T_{O_i},\,^{O_i}T_G,\,^W\!T_{O_f})\), grasps in world at pick and place are formed as:
\(^{W}\!T_{G_i} \! =\!  ^W\!T_{O_i}\,^{O_i}T_G\) and \(^{W}\!T_{G_f} \! =\! ^W\!T_{O_f}\,^{O_i}T_G\). Then for each phase \(x\in\{i,f\}\) (pick \(i\), place \(f\)), three defined waypoints that are based in world-frame:
\begin{itemize}
    \item \textbf{Grasp} \(G(x)\): grasp pose at the object for phase \(x\).
    \item \textbf{Pre-grasp} \(P(x)\): translated from \(G(x)\) by a fixed offset along the tool’s approach axis \emph{away from the object}
    (i.e., along \(-Z_{\text{tool}}\) if \(+Z_{\text{tool}}\) is the approach direction).
    \item \textbf{Lift} \(L(x)\): translated from \(G(x)\) by a fixed height \(\Delta z\) along world \(+z\) (vertical clearance).
\end{itemize}

The canonical pick-and–place sequence is:
\[
P(i)\ \rightarrow\ G(i)\ \rightarrow\ L(i)\ \rightarrow\ L(f)\ \rightarrow\ P(f)\ \rightarrow\ G(f).
\]
Motion between successive waypoints uses straight-line Cartesian interpolation with smooth orientation interpolation (or a motion plan that respects the same waypoints). Because $P(x)$ and 
$L(x)$ are defined by fixed offsets from $G(x)$, once $(^{W}\!T_{O_i},\,^{O_i}\!T_G,\,^{W}\!T_{O_f})$ and the offset parameters are given, all waypoint poses can uniquely be determined and do not change across runs or planners.

\subsection{Ground Truth Labels}

Two binary properties for each candidate are defined:

\begin{enumerate}
\item \textbf{Path-wise IK feasibility} ($y_{\mathrm{IK}}\!\in\!\{0,1\}$)\;:\;
$y_{\mathrm{IK}}=1$ iff IK solutions exist (under joint limits) at all template waypoints and the interpolated joint segments between them respect joint limits. (Collisions excluded.)

\item \textbf{Collision-free label} ($y_{\mathrm{COL}}\in\{0,1\}$)\; :
$y_{\mathrm{COL}}=1$ iff the sequence
$L(i)\!\rightarrow\!L(f)\!\rightarrow\!P(f)\!\rightarrow\!G(f)$
is free of robot–self, robot–environment, and object–environment collisions. Here, Collisions are checked on one fixed IK branch: pick one valid IK, for each later waypoint, start from the previous joint solution and choose the smallest joint-change solution. Thus $y_{COL}=1$ denotes collision-free \emph{under this policy}.
Intended gripper–object contact is allowed, and table contact with the object at the final release is allowed.
\end{enumerate}



\begin{figure*}[!t]
  \centering
  \includegraphics[
    width=0.8\textwidth,
    trim=40mm 20mm 30mm 20mm, 
    clip
  ]{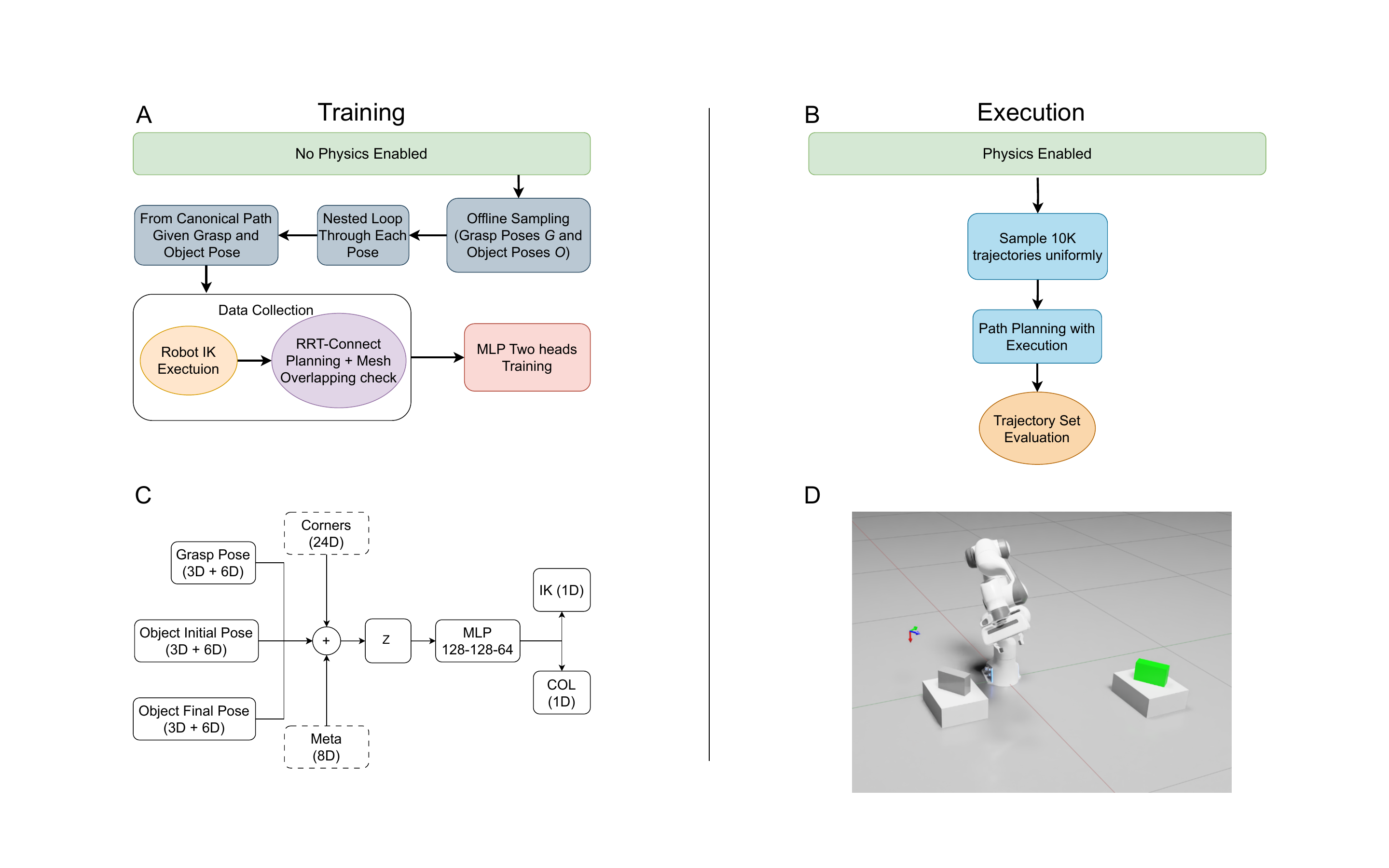}
\caption{(A) Training flow, no physics. Sample object and grasp poses, build the canonical P–G–L path, compute path-wise IK and path-sweep collision labels, and train a dual-output MLP. 
(B) Execution in Sim, physics enabled. Score held-out labels and run 10k executed-sim trajectories by planning with RRT-Connect and executing in PhysX to compare predictions to outcomes. 
(C) A dual-output MLP scores the triplet $(^{W}\!T_{O_i},\,^{O_i}\!T_G,\,^{W}\!T_{O_f})$. Inputs are three pose blocks: grasp in the initial object frame $\phi_G=[t_G,\,r(R_G)]$, and object poses at pick/place in world $\phi_i=[t_i,\,r(R_i)]$, $\phi_f=[t_f,\, r(R_f)]$,$r(\cdot)$ is a 6D rotation (first two matrix columns). There are two optional descriptors for ablation studies: final-pose corners' positions ($8{\times}3$) in world frame and a grasp meta vector $\phi_m=[u_{\mathrm{frac}},v_{\mathrm{frac}},onehot(face)]$ ($u_{frac}, v_{frac} \in [0,1]$, they are fractions along the face that will be grasped and $onehot(face)$ indicates which face is grasped). The available blocks are concatenated into a fused feature embedding $z$ and processed by a trunk MLP (128–128–64) with two heads predicting IK and collision scores.
(D) An example scene of actual trajectory. 7-DoF arm moving a single cuboid between pedestals using the canonical pre-grasp (P), grasp (G), lift (L) template. The grey box represents the actual object, the green box is the target pose for visualization only.}
  \label{fig:system_picture}
\end{figure*}

\section{Approach}
\noindent This section outlines a four-stage pipeline (generation $\rightarrow$ training $\rightarrow$ executed-sim validation $\rightarrow$ deployment), see Fig.~\ref{fig:system_picture} for a visual summary.

\begin{itemize}

    \item \textbf{Data synthesis:} Paired pick/place episodes are synthesized from the P/G/L template, and two labels are assigned per candidate: (i) path-wise IK feasibility—IK solvable at the required template waypoints—and (ii) a path-aware collision flag from mesh-overlap sweeps along the nominal pick–place path $L(i)\!\rightarrow\!L(f)\!\rightarrow\!P(f)\!\rightarrow\!G(f)$.

    \item  \textbf{Supervised training:} A dual-output MLP encodes the triplet $(^W\!T_{O_i},\,^{O_i}T_G,\,^W\!T_{O_f})$ (optionally with descriptors) and predicts IK and collision feasibility scores.
    \item \textbf{Executed-sim validation:} On a held-out set of 10k candidates, the nominal P/G/L trajectory $\Gamma(T_{O_i},\, G_{O_i},\, T_{O_f})$ is instantiated. A path-wise IK gate is applied over the template waypoints, planning is done with RRT-Connect, execution runs in PhysX, and the model’s feasibility scores are compared to planner success and collisions observed along the executed path.

    \item \textbf{Intended deployment:} At runtime, given $(T_{O_i}, T_{O_f})$, the precomputed grasp set $\{G_{O_i}\}$ is scored to produce $(s_{\mathrm{IK}}, s_{\mathrm{COL}})$, candidates are ranked, and only the top selection(s) are executed.

\end{itemize}

\begin{algorithm}[t]
\caption{GenerateGrasps}
\label{alg:one}
\KwIn{Object pose (identity here); lattice sets $\mathcal U,\mathcal V$; tool offsets $\mathcal R$; contact offset $d_{\mathrm{off}}$}
\KwOut{Candidates $\mathcal G$ with contact/hand/tool/meta}
$\mathcal G \gets \emptyset$\;
\ForEach{$f \in \{+X,-X,+Y,-Y,+Z,-Z\}$}{
  compute face-frame unit vectors: $L$ (long), $S$ (short), $n$ (normal)\;
  set approach direction $a \gets -\,n$\tcp*[r]{gripper approaches opposite the outward normal}
  \ForEach{$u \in \mathcal U$}{
    \ForEach{$v \in \mathcal V$}{
      $p_{\mathrm{loc}} \gets \textsc{FaceCenter}(f) \;+\; u\,L \;+\; v\,S \;+\; d_{\mathrm{off}}\,n$\tcp*[r]{contact in object frame}
      $R_{\mathrm{base}} \gets \textsc{BaseTool}(a,S)$\tcp*[r]{approach $\parallel a$, closing axis $\parallel S$}
      \ForEach{$r \in \mathcal R$}{
        $R_{\mathrm{tool}} \gets \textsc{ApplyRotations}(R_{\mathrm{base}}, r)$\;
        $p_W \gets {}^W\!R_{O}\,p_{\mathrm{loc}} + {}^W\!t_{O}$\tcp*[r]{contact position in world}
        $R_W \gets {}^W\!R_{O}\,R_{\mathrm{tool}}$\tcp*[r]{tool orientation in world}
        $H_W \gets \textsc{WristPose}(p_W, R_W)$\tcp*[r]{hand/wrist pose from contact \& tool}
        append $\big(p_W, H_W, R_W,\,(u,v,onehot(face))\big)$ to $\mathcal G$\;
      }
    }
  }
}
\Return $\mathcal G$\;
\end{algorithm}

\subsection{\textbf{Data synthesis}}\label{sec:data_synthesization}
To build the dataset, object world poses and associated grasp candidates are synthesized, then a nominal P/G/L path is evaluated to assign labels. Object poses are sampled by pairing positions across the reachable workspace with stable orientations (support face and in-plane spin). Grasp candidates are defined in the object frame and mapped to the world at each sampled pose. During synthesis, physics is disabled and fast geometric checks are used to compute path-wise IK feasibility and transit-collision labels. Full physics execution is reserved for held-out evaluation.

\subsubsection{Object pose and pedestal generation}

\begin{figure}[t!]
    \centering  \includegraphics[width=0.5\columnwidth]{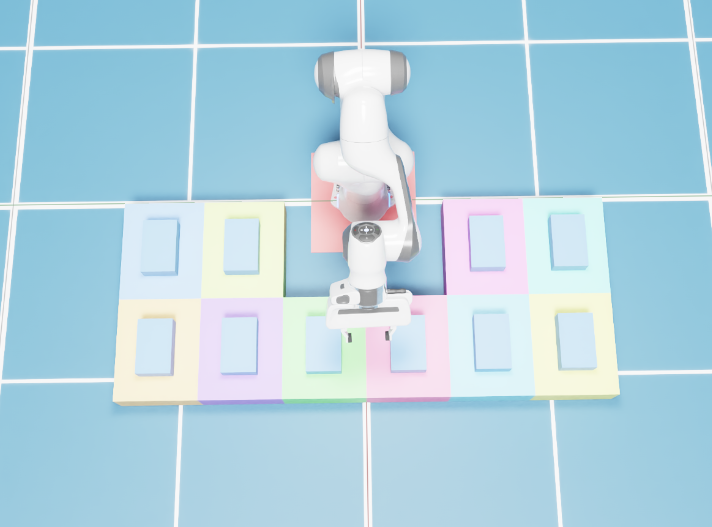}
    \caption{The setup shown in the picture is how pedestals are placed, and the robot is placed at the origin }
    \label{fig:pedestal_poses}
\end{figure}

Object poses ${}^W\!T_{O}=(t_{O},\,R_{O})$ are sampled by pairing a support-face orientation with a pedestal center. Pedestals are axis-aligned rectangles arranged as the largest zero-gap grid that fits inside a rectangular robot workspace safe span. The robot is placed at the top center of the workspace. Any grid cell whose pedestal footprint overlaps the robot base is discarded. Each admissible pedestal center $(x_k,y_\ell)$ defines a placement where the object shares the same planar position $(x,y)$ as the pedestal, and the overall layout is shown in Fig.~\ref{fig:pedestal_poses}.

\begin{figure}[t!]
    \centering
    \includegraphics[width=0.4\linewidth]{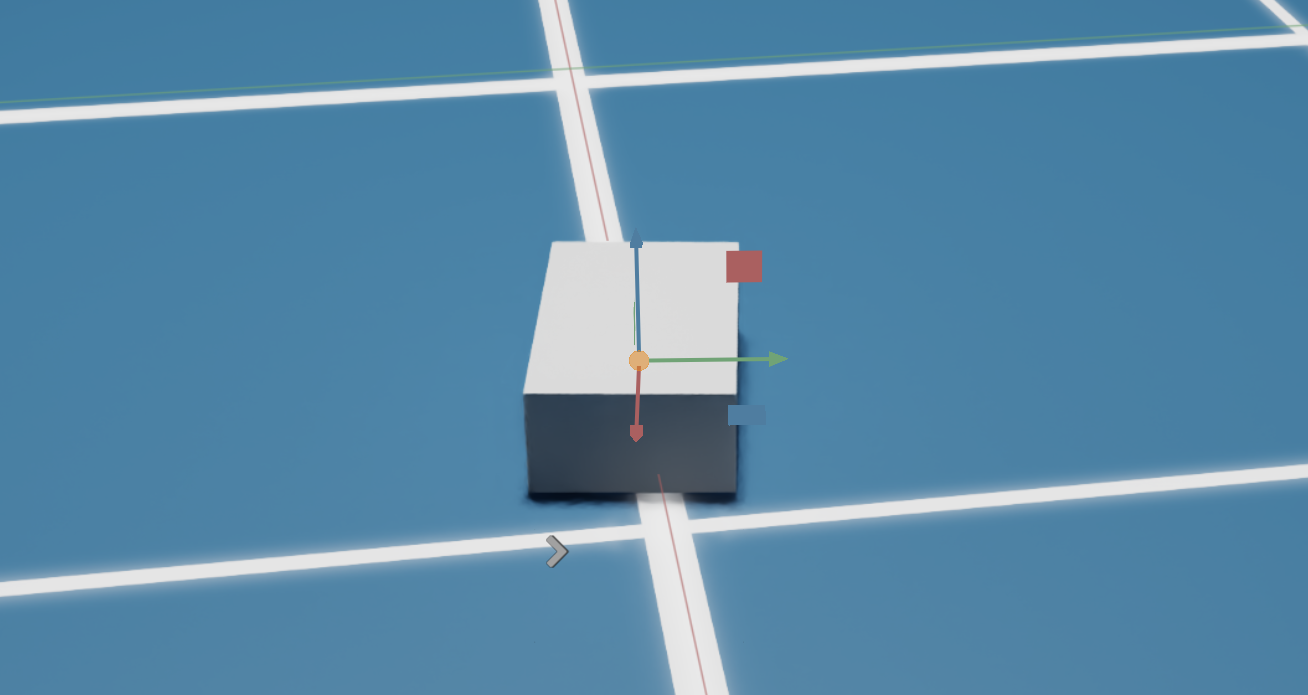}
    \includegraphics[width=0.4\linewidth]{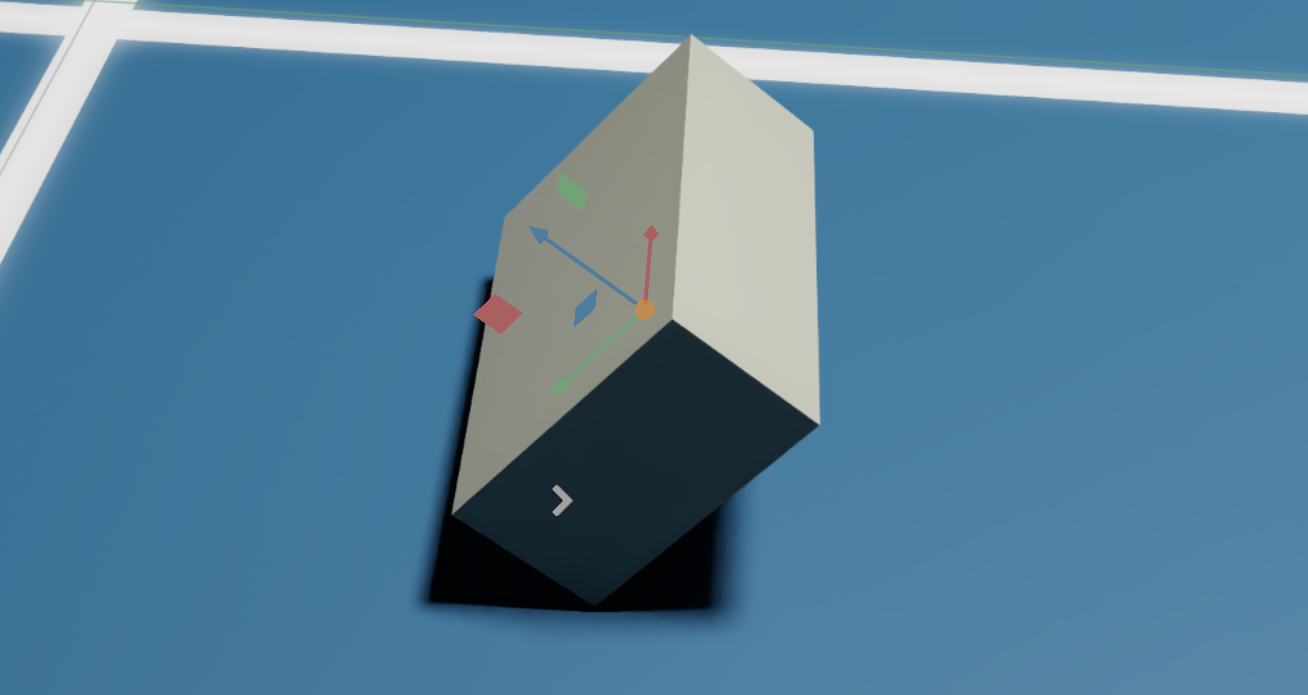}
    \caption{Comparison of object orientations. \textbf{Left:} a stable pose in which the object can rest on a surface (feasible placement). \textbf{Right:} an unstable orientation that the object cannot maintain (infeasible pose).}
    \label{fig:feasible_comparison}
\end{figure}

Orientations are restricted to statically stable (“feasible”) poses, meaning the object would rest without tipping when physics is enabled (Fig.~\ref{fig:feasible_comparison}). Sampling proceeds by first choosing a support face $s\!\in\!\{\text{top},\text{bottom},\text{left},\text{right},\text{front},\text{back}\}$, a face-consistent base pose is then fixed in which $s$ lies on the horizontal support plane, and a spin angle is drawn uniformly over $[0,2\pi)$ about the support-face normal. This keeps the same face down while covering the full in-plane rotation range.

\subsubsection{Grasp pose generator}\label{sec:grasp_pose_gen}

\begin{figure}[H]
 \centering  \includegraphics[width=0.9\columnwidth]{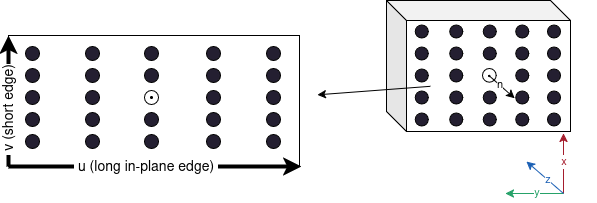}
\caption{Grasp parameterization (single composite view). 
\textit{Left:} in-plane lattice on a side face — $u$ runs along the longer in-plane edge and $v$ along the shorter; black dots are candidate contacts and the hollow dot marks $(u{=}0.5, v{=}0.5)$. 
\textit{Right:} the corresponding grasp on the object — red: gripper-closing axis, green: binormal axis, blue: approach axis; the face normal $n$ points opposite the local $z$ shown (i.e., $n = -\hat z$ for that face)}
\label{fig:grasp_visualization}
\end{figure}

For each face $\{\pm X, \pm Y, \pm Z\}$, the in-plane axes (long and short edges), the outward normal, and the approach direction (into the face) are defined. The lattice uses parameters $(u,v)$ with $u$ along the longer in-plane axis and $v$ along the shorter, and each $(u,v)$ yields a local grasp point slightly offset outward along the face normal. Around every grasp point, a small fixed set of local rotations (yaw/tilt/roll) in the tool frame is enumerated to cover approach variability. Figure~\ref{fig:grasp_visualization} illustrates how $u$ and $v$ parameterize a generated grasp pose. Cases that are infeasible to grasp—e.g., object dimensions exceeding the gripper’s maximum width—are filtered out. For each grasp candidate, the grasp pose is stored in the object frame as ${}^{O}\!T_{G}$. For any object world pose ${}^{W}\!T_{O}$, the world grasp pose is ${}^{W}\!T_{G} = {}^{W}\!T_{O}\,{}^{O}T_{G}$.

Beyond grasp poses, three simple \emph{meta features} are used:
\[
m = \big[u_{\mathrm{frac}},\ v_{\mathrm{frac}},\ \mathrm{onehot}(\text{face})\big].
\]
Here $u_{\mathrm{frac}}\in[0,1]$ is the fraction along the face’s \emph{long} in-plane edge (0 = one long-side edge, 0.5 = center, 1 = the opposite long-side edge),
$v_{\mathrm{frac}}\in[0,1]$ is the fraction along the \emph{short} in-plane edge,
and $\mathrm{onehot}(\text{face})$ indicates which face is grasped. These features record \emph{where} on the face the grasp sits and \emph{which} face is used—information that helps the model reason about reachability and clearance near edges. The implementation pseudocode is given in Algorithm~\ref{alg:one}.

\subsubsection{Raw Episode Collection (P/G/L primitives)}\label{sec:raw_collection}
Each raw episode is logged, indexed by $(p,g,o)$ where $p$ is an object position sample, $g$ is a grasp ID, and $o$ is an object orientation sample. For each position $p$, orientation $o$, and grasp $g$, the world-frame grasp pose and its primitives are defined as
\[
\begin{aligned}
G(x) &:= \mathrm{Grasp}(x) = {}^{W}\!T_{O_x}\,{}^{O_x}\!T_{G}, \quad x\in\{i,f\},\\
P(x) &:= \mathrm{Approach}\big(G(x)\big),\\
L(x) &:= \mathrm{Lift}\big(G(x)\big).
\end{aligned}
\]

The path-sweep flag along $P(x)\!\to\!G(x)\!\to\!L(x)$ is computed via discrete IK feasibility checks and collision tests. Motions between the fixed waypoints (e.g., $P\!\rightarrow\!G\!\rightarrow\!L$) are planned with RRT-Connect. If the planner finds a path for the required segments, the case is marked as IK-feasible, otherwise, it is not. Collision status is checked by mesh–mesh overlap tests between robot links and the static environment (pedestals and ground) along the planned motion.

\subsection{\textbf{Training Pairs}}\label{sec:pairing}
Raw episodes are single–grasp trials at specific object world poses, generated with physics \emph{disabled}. The same generator produces episodes that are later designated as \emph{pick} or \emph{place}. A training pair is formed by matching a pick episode $(p_1, g, o_1, \ldots)$ with a place episode $(p_2, g, o_2, \ldots)$ that shares the same object–frame grasp ID $g$ and differs in pose, subject to route and coverage constraints. Each valid match instantiates a nominal P/G/L transit from $(p_1,o_1)$ to $(p_2,o_2)$, from which labels are derived.

\subsubsection{Constraints and binning.}
Two episode sets are generated with the same procedure but different roles: a \emph{pick} set and a \emph{place} set, where each episode is a triple $(p,g,o)$ of planar position $p$, grasp ID $g$, and orientation $o$. Only \emph{valid} pick episodes are retained after basic mechanical checks are applied (e.g., jaw opening within limits, unobstructed support).

For each valid pick $(p_1,g,o_1)$, the candidate pool for pairing is all place episodes that share $g$ but differ in pose. This pool is filtered by removing the identical triple, restricting to an allowed list of position-to-position edges so the transit is meaningful (route constraint), and requiring the same basic endpoint checks at the place pose.

To obtain diverse yet controlled coverage, each surviving place candidate $(p_2,g,o_2)$ is tagged by two quantities: translation magnitude $d=\|p_2-p_1\|$, binned from \emph{near} to \emph{far} via fixed distance thresholds, and orientation change $\theta$ between $o_1$ and $o_2$ (the geodesic/quaternion angle), binned from \emph{small} to \emph{large}. Selection then iterates over the distance$\times$angle bins. In each pass, one representative is taken from each nonempty bin and the rest are skipped so no bin dominates. Reuse of any exact place episode $(p_2,g,o_2)$ across different picks is capped by $R_{\mathrm{cap}}$. Bins are cycled until either $K$ pairs are formed for the current pick or the pool is exhausted (Algorithm~\ref{alg:two}).

\begin{algorithm}[t]
\caption{Pair episodes}
\label{alg:two}
\DontPrintSemicolon
\ForEach{pick $(p,g,o)$ in \textsc{PICK}}{
  \If{\textbf{not} \textsc{ValidPickBasic}$(p,g,o)$}{\textbf{continue}}
  pool $\gets \{(p_2,g,o_2)\in\textsc{PLACE}:\ (p_2\neq p)\ \textbf{or}\ (o_2\neq o)\}$\;
  cands $\gets \big\{(p_2,g,o_2)\in\text{pool}:\ (p\!\to\!p_2)\in \textsc{ValidPlaceBasic}(p_2,g,o_2)\big\}$\;
  buckets $\gets$ group \textit{cands} by $\big(\textsc{DistBin}(\|c[p]-c[p_2]\|),\ \textsc{AngBin}(\textsc{QuatAngle}(o,o_2))\big)$\;
  pairs $\gets [\,]$\;
  \ForEach{$key$ in \textsc{Ordered}(buckets)}{
    \tcp{$e=(p_2,g,o_2)$ is a place episode}
    \ForEach{$e=(p_2,g,o_2)$ in buckets[$key$]}{
      \If{$used[(p_2,g,o_2)] < R_{\mathrm{cap}}$}{
        pairs.append$(((p,g,o),(p_2,g,o_2)))$;\quad
        $used[(p_2,g,o_2)] \leftarrow used[(p_2,g,o_2)] + 1$;\quad
        \textbf{break} 
      }
    }
    \If{$|$pairs$| = K$}{\textbf{break}}
  }
  \textbf{output} pairs for this pick\;
}
\end{algorithm}

\subsubsection{Labels}\label{sec:labels}
For each accepted \emph{place} episode (physics disabled), three raw flags are computed:
\[
\texttt{ik},\ \texttt{pt},\ \texttt{col} \in \{0,1\}.
\]
\begin{itemize}
  \item \texttt{ik}: \emph{path-wise IK feasibility} over the place sweep 
  $\mathcal{S}(t)$ from $P(t)$ to $G(t)$—i.e., IK solutions exist at all sampled waypoints on $\mathcal{S}(t)$ and the joint-space interpolation between successive IK solutions respects joint limits. (No collision checks inside this IK test.)
  \item \texttt{pt}: \emph{path collision} along the same sweep $\mathcal{S}(t)$, via mesh-overlap sweeps (1 if any collision is detected along the path).
  \item \texttt{col}: \emph{endpoint collision} at the place grasp $G(t)$ (1 if in collision).
\end{itemize}

then in the training targets, labels are defined as
\[
\begin{aligned}
y_{\mathrm{IK}}  &:= \mathbf{1}\{\texttt{ik}=1\},\\[3pt]
y_{\mathrm{COL}} &:= \mathbf{1}\big\{(\texttt{ik}\wedge \texttt{pt}) \ \vee\ (\neg\,\texttt{ik}\wedge \texttt{col})\big\}.
\end{aligned}
\]
Thus, when the place sweep is kinematically solvable (\texttt{ik}=1), $y_{\mathrm{COL}}$ uses the \emph{path} collision flag, otherwise it falls back to the \emph{endpoint} collision at $G(t)$.


\subsection{\textbf{Model Architecture}}\label{sec:model}
Each triplet $(^W\!T_{O_i},\,^{O_i}T_G,\,^W\!T_{O_f})$ is scored with a dual-output MLP. 
Inputs are encoded in the frames that are already used: the grasp in the initial object frame, and the object poses in world frame at pick and place.
Let $^W\!T_{O_i}=(t_i,R_i)$, $^W\!T_{O_f}=(t_f,R_f)$, and $^{O_i}T_G=(t_G,R_G)$,
three feature blocks can be formed as:
\[
\phi_G=\begin{bmatrix} t_G \\ r(R_G) \end{bmatrix},\qquad
\phi_i=\begin{bmatrix} t_i \\ r(R_i) \end{bmatrix},\qquad
\phi_f=\begin{bmatrix} t_f \\ r(R_f) \end{bmatrix}.
\]

where $r(\cdot)$ denotes a compact, continuous rotation encoding.
(Positions are normalized by fixed scales—grasp by box size, world by workspace—and rotations use a 6D vector given by the first two columns of the rotation matrix, providing a continuous parameterization\cite{Zhou_2019_CVPR}), the model architecture can be seen from Fig. \ref{fig:system_picture}C

When enabled, two optional descriptors are appended. First, the eight target corners in world \(C_f(\mathbf d)\in\mathbb{R}^{8\times3}\) derived from \(^W\!T_{O_f}\), these summarize the object’s extent and orientation at the target and are mapped by a small MLP and LayerNorm to a vector \(\phi_C\). Second, a grasp meta vector \(m=[u_{\mathrm{frac}},\,v_{\mathrm{frac}},\,onehot(face)]\) indicating where the grasp sits on the face and which side face is used and this is mapped by a small MLP to \(\phi_m\). The available blocks are concatenated,
\[
z=\mathrm{concat}\big(\phi_G,\ \phi_i,\ \phi_f,\ [\phi_C],\ [\phi_m]\big).
\]
The vector $z$ is fed to a trunk MLP and two output heads produce the scores $(s_{\mathrm{IK}}, s_{\mathrm{COL}})$.

Given labels \((y_{\mathrm{IK}},y_{\mathrm{COL}})\in\{0,1\}^2\), the training loss is
\[
\mathcal L=\mathrm{BCE}(s_{\mathrm{IK}},y_{\mathrm{IK}})\;+\;\mathrm{BCE}(s_{\mathrm{COL}},y_{\mathrm{COL}}),
\]
optionally using per-class weights for imbalanced data.
For efficiency, encodings for \((^W\!T_{O_i},\,^{O_i}T_G,\,^W\!T_{O_f})\) are cached, while corners and meta are computed on the fly.

\section{Experiments and Results}
\subsection{Simulation Setup}
The data collection was conducted in the NVIDIA Isaac Sim virtual environment. The setup consists of a Franka Emika Panda 7-DoF robotic arm and a cuboid object with dimensions $14.3~\text{cm}\times 9.15~\text{cm}\times 5.1~\text{cm}$. The object was placed at different positions on top of available cuboid pedestals.

\subsection{Training and evaluation}
A dual-output MLP was trained (Sec.~\ref{sec:model}) with AdamW (lr $=1\!\times\!10^{-3}$, weight decay $=3\!\times\!10^{-4}$), batch size $4096$, hidden size $64$, dropout $0.10$, global gradient–norm clipping $2.0$, a $5$–epoch warmup then decay, and early stopping (patience $10$). The loss is the sum of two class–weighted BCE terms (IK, Collision), with class weights computed from training–set label frequencies. The optional \emph{Corners} and \emph{Meta} features are enabled by default and ablated alongside alternative input formats.

For the evaluations, evaluations were in two settings: \begin{itemize}
    \item A held-out \emph{label-only} split (no physics) used for per-head Accuracy
    \item An \emph{executed-simulation} set used as external validation of transfer
\end{itemize}
\textbf{Executed-simulation protocol.} From the held-out pool, $10{,}000$ candidates $(^W\!T_{O_i},\,{}^{O_i}\!T_G,\,^W\!T_{O_f})$ were sub-sampled, covering pedestal-distance and orientation-change bins. For each candidate:
\begin{enumerate}
\item Instantiate the fixed waypoint sequence $P(i)\!\rightarrow\!G(i)\!\rightarrow\!L(i)\!\rightarrow\!P(f)\!\rightarrow\!G(f)\!\rightarrow\!L(f)$ (P/G/L template).
\item Apply a \emph{path-wise IK gate} over this sequence. Solve IK at all required waypoints and ensure the joint-space interpolation between successive IK solutions respects joint limits. If unsatisfied, label as IK-infeasible and skip planning.
\item If the gate passes, plan with RRT-Connect using fixed parameters and timeouts and execute in PhysX with the object not artificially attached. Record planner success and collisions observed along the executed path.
\end{enumerate}




\subsection{Ablation Studies}\label{subsec:ablations}
Object-pose and grasp-pose representation choices are studied for their effect on performance on both the label-only set and the trajectory set. Across all runs, data splits, seeds, optimizer, and decision thresholds are fixed. The base input is the triplet $(^W\!T_{O_i},\,^{O_i}T_G,\,^W\!T_{O_f})$ encoded as three pose blocks (pick object pose, grasp pose in the initial object frame, and target object pose). Four options are toggled, and the inputs that differ in each case are stated.

\begin{itemize}

\item \textbf{Frames (Abs vs.\ Rel poses):}
Absolute world poses are compared to relative pick$\rightarrow$place poses. In the \textbf{Abs} variant, the object-pose blocks use absolute world poses \((t_i, R_i)\) and \((t_f, R_f)\) from \(T_{O_i}\) and \(T_{O_f}\). In the \textbf{Rel} variant, the place block is represented by the relative pose \((t_\Delta, R_\Delta)\), where \(t_\Delta = t_f - t_i\) (normalized by table extents) and \(R_\Delta = R_f R_i^\top\). Both variants use the same 6D rotation encoding given by the first two columns of the rotation matrix.

  \item \textbf{Meta features (ON vs.\ OFF):}
 Meta adds a small vector that encodes where the grasp sits on the face and which face is used: $m=[u_{\text{frac}},\,v_{\text{frac}},\,\text{onehot}(\text{face})]$ (see Sec.~\ref{sec:grasp_pose_gen}). When \textbf{ON}, an MLP embedding of $m$ is appended to the pose features. When \textbf{OFF}, $m$ is omitted and no replacement is introduced.

  \item \textbf{Corners (Crn on/off):} Cuboid–corner features inspired by \cite{jeon2023learning} are used to improve generalization by exposing local target geometry and clearance. When \textbf{on}, an embedding of the eight world-frame target corners $C_f(\mathbf d)$, computed from $T_{O_f}$ (small MLP + LayerNorm), is appended. When \textbf{off}, the corner block is omitted with no substitute and the network still receives object pose via $T_{O_i}$ and $T_{O_f}$. Corners are expressed in the world frame and the ordering is given in Fig.~\ref{fig:vertices}.

  \begin{figure}[h]
    \centering  \includegraphics[width=0.7\columnwidth]{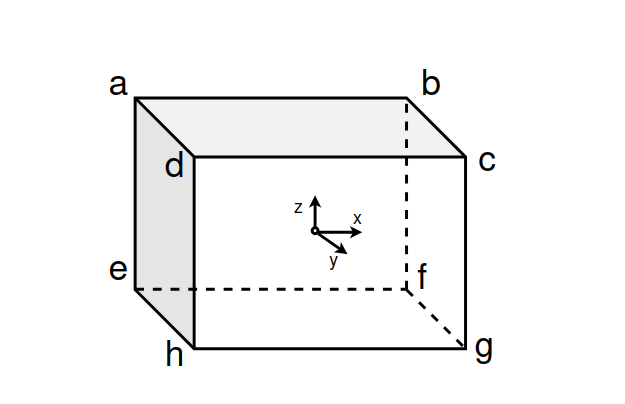}
    \caption{Object frame at the geometric center. Corners are listed by face then clockwise: the $-X$ face is $[e,a,d,h]$ and the $+X$ face is $[f,b,c,g]$, giving the sequence $e\!\rightarrow\!a\!\rightarrow\!d\!\rightarrow\!h\!\rightarrow\!f\!\rightarrow\!b\!\rightarrow\!c\!\rightarrow\!g$.}

    \label{fig:vertices}
\end{figure}
\item 
\textbf{Grasp pose (Gpose: Pick vs.\ Place).} In \textbf{Pick}, the grasp at the pick stage in the pick frame, ${}^{O_i}T_{G_{i}}$ (tool pose relative to $O_i$), is used. In \textbf{Place}, the same grasp carried to the placement stage but still expressed in the pick frame, i.e., ${}^{O_i}T_{G_{f}}$, is used. This parameter changes only the grasp features, and place features are controlled separately. Keeping a single reference frame removes absolute world pose and tests whether representing the grasp “as it will be at place” improves IK/collision prediction for the fixed template.

\end{itemize}

\subsection{Baseline–Model Comparison}
This comparison isolates the value of learned grasp prioritization under identical motion generation. Two methods are evaluated in 100 trials sampled from the trajectory set:
\begin{itemize}
    \item \textbf{Heuristic:} From the surviving pool, the Baseline chooses exactly one grasp \emph{at random} from each of the three available faces, yielding three attempts per trial.
    \item \textbf{Model selection:} From the same surviving pool, the Model ranks candidates by its predicted feasibility score and selects the top three grasps (three attempts).
\end{itemize}

A random per-face baseline is compared to the model under identical scene and planning conditions. Each trial reuses an initial pose $T_{O_i}$ and a target pose $T_{O_f}$ from the prior trajectories. A shared grasp pool is built in the object frame with candidates balanced across the surfaces $\{+X,-X,+Y,-Y\}$. Filtering is applied solely by the initial object pose: the support face at $T_{O_i}$ is removed and the $\{+Z,-Z\}$ faces are excluded (incompatible with the gripper width) while the remaining faces are retained. No filtering is applied based on the placement $T_{O_f}$.

The baseline draws one grasp at random from each of the three surviving faces, giving three attempts per trial. If a face has no surviving candidates, the missing grasp is drawn uniformly from the remaining faces to keep three attempts. The model ranks the same surviving pool by its feasibility scores and selects the top three grasps.

Both methods execute the same motion template,
\(P_1 \rightarrow C_1 \rightarrow L_1\), transit, \(L_2 \rightarrow P_2 \rightarrow C_2\), followed by a short retreat. Planning parameters, controller settings, timeouts, and collision criteria are identical. Differences in success or failure rates stem solely from grasp selection under identical planning, underscoring that learned ranking—not physics or dynamics—drives the improvement.

\begin{table}[H]
\small
\centering
\caption{Executed-sim runtime (same $\mathcal G$, IK gate, planner, budget $B$).}
\label{tab:baseline_runtime}
\begin{tabular}{lcccc}
\toprule
Method & Succ@1 & Succ@2 &  Succ@3 & Fail\\
\midrule
B1: Brute+IK gate & 47 &  15 & \textbf{12} &  26 \\
Ours (Main)       & \textbf{53} &  \textbf{30} & 4 & \textbf{13}  \\
\bottomrule
\end{tabular}

\vspace{2pt}
\raggedright\footnotesize
Success@K: solved within $K$ planner calls, Calls/Time to first success. Also report planned-path collision rate and $(e_t,e_R)$ for successes.
\end{table}

\begin{table*}[!t]
\small
\setlength{\tabcolsep}{3pt}
\centering
\caption{Ablations with held-out and executed-sim accuracies (same splits/seeds): \emph{Label-set:} evaluates per-head classification accuracy (IK, COL) on the route-disjoint test split using the labels (no physics). \emph{Trajectory-set:} evaluates the same heads on the $10\mathrm{k}$ executed trajectories, where IK is measured by planner success (with the path-wise IK gate) and COL by path collisions observed during execution. \textbf{Frms} (object pose): \textbf{Abs} feeds world-absolute $(T_{O_i},T_{O_f})$ and \textbf{Rel} feeds the relative pose $\Delta=T_{O_i}^{-1}T_{O_f}$ as $[t_\Delta,\,6\mathrm{D}(R_\Delta)]$. 
\textbf{Meta}: append $m=[u_{\mathrm{frac}},v_{\mathrm{frac}},onehot(face)]$ (ON) or omit (OFF). 
\textbf{Crn}: append the eight target corners $C_f(\mathbf d)$ from $T_{O_f}$ (ON) or omit (OFF). 
\textbf{Gpose} (grasp): \textbf{Pick} uses $G_O$ (grasp at pick in $O_i$); \textbf{Place} uses $G_\Delta=\Delta\,G_O$ (grasp at place, expressed in $O_i$). 
Metrics: \textbf{IK} is kinematic head and \textbf{COL} is collision head. Main = Frms:Abs, Meta:ON, Crn:ON, Gpose:Pick}
\label{tab:abl_combo_split}
\begin{tabular}{lcccccccccc}
\toprule
& & & & & & \multicolumn{2}{c}{Label set Acc} & \multicolumn{2}{c}{Trajectory set Acc} \\
\cmidrule(lr){7-8}\cmidrule(lr){9-10}
ID & Frms & Meta & Crn & Gpose & Inputs (dims) & IK & COL & IK & COL \\
\midrule
Main            & Abs  & \checkmark & \checkmark & Pick  & Oi7+Gp7+Of7 + M8 + C24 $=\mathbf{53}$ & 0.982 & 0.968 & 0.899 & 0.928 \\
A1 (--Crn)      & Abs  & \checkmark & $\times$    & Pick  & Oi7+Gp7+Of7 + M8 $=\mathbf{29}$       & 0.980 & 0.967 & 0.898 & 0.929 \\
A2 (Rel)        & Rel  & \checkmark & \checkmark  & Pick  & Rel7+Gp7 + M8 + C24 $=\mathbf{46}$     & 0.983 & 0.967 & 0.899 & 0.929 \\
A3 (--Meta)     & Abs  & $\times$    & \checkmark  & Pick  & Oi7+Gp7+Of7 + C24 $=\mathbf{45}$       & \textbf{0.983} & \textbf{0.969} & \textbf{0.899} & \textbf{0.930} \\
A4 (Gpose)      & Abs  & \checkmark  & \checkmark  & Place & Oi7+Gpl7+Of7 + M8 + C24 $=\mathbf{53}$ & 0.982 & 0.968 & 0.804 & 0.838 \\
\bottomrule
\end{tabular}

\vspace{2pt}
\emph{Dims shorthand:} Oi7 = object pick pose (7D), Of7 = object place pose (7D), Rel7 = relative pose $\Delta$ (7D), Gp7 = grasp at pick in $O_i$ (7D), Gpl7 = grasp at place expressed in $O_i$ (7D), M8 = meta (2 fracs + 6-face one-hot), C24 = 8 corners $\times$ 3D.
\end{table*}

\section{Discussion}

\subsection*{Path-aware labeling and limitations}
Endpoint-only checks are not sufficient for reliable supervision, and the current scope—single cuboid and side-face grasps—further constrains generality but enables controlled analysis. In the data, trials are repeatedly observed where the final place pose is IK-feasible and free of endpoint collisions, yet the physics-enabled trajectory still collides during one of the primitive segments. Typical failure modes include grasp during $P_1\!\rightarrow\!G_1$ as the gripper enters between the object and the pedestal, scraping during $G_1\!\rightarrow\!L_1$ when clearances are small, and self or environment collisions during the long $L_1\!\rightarrow\!L_2$ transit. If the label depends only on the final pose then these become false positives that the model cannot learn to avoid. Adding a path sweep to the collision label directly targets these cases and produces scores that match execution much better. In practice, a sweep is run along all primitive segments of the template with discrete samples in task space, and the label is set to negative if any contact is detected. This makes the supervision more conservative. A further improvement is to expand the path-aware notion on the IK side as well by checking IK at intermediate samples and by penalizing near-singular configurations, for example using a Jacobian condition metric.

\subsection*{Feasibility versus planning difficulty}
Placement execution ultimately depends on a motion planner to produce a collision-free trajectory. In experiments with an RRT-based planner, failures were observed even when a path existed (especially in cluttered or highly constrained scenes). In these cases the model correctly flagged the candidate as feasible (path-wise IK solvable and transit collision-free under the checks), yet execution failed because the planner’s search did not uncover a valid path within the budget.

This gap indicates that the predictor estimates geometric feasibility, not search complexity. In practice, higher feasibility scores generally track execution success, but runs can still fail when the planner times out or gets stuck (e.g., tight clearances, clutter, or a small planning budget). A simple diagnostic to expose this distinction is to report both the model’s feasibility score and whether the planner succeeded. Adding this would clarify whether errors stem from geometry (labels and model) or from planning (search) and can be left to future iterations.

\subsection*{Ablations and baseline approach}
Table~\ref{tab:abl_combo_split} shows four controlled variants. The trends are compact:
\begin{itemize}
\item \textbf{Corners (A1).} Removing target corners changes both heads by at most $\le\!0.002$ on held-out and leaves executed accuracy essentially unchanged. For a single known cuboid, $(^W\!T_{O_i},\,{}^{O_i}\!T_G,\,^W\!T_{O_f})$ already encodes target geometry.
\item \textbf{Relative Frames (A2).} Using $^W\!T_{O_i},\,^W\!T_{O_f}$ yields a small, consistent lift on executed-sim (COL $\approx\!+0.0014$, IK $\approx\!+0.0001$). Best practice is to \emph{retain absolute world context and add} relative cues.
\item \textbf{No meta (A3).} Dropping grasp meta gives the strongest overall scores (executed: COL $0.9296$, IK $0.8994$), suggesting pose features subsume the meta bits and avoid weak correlations.
\item \textbf{Grasp pose in Placement phase (A4).} Looks fine offline but drops sharply in executed-sim (IK $-9.6$ pts, COL $-9.0$ pts vs.\ Main). World-frame context is critical for real path feasibility.
\end{itemize}
Overall, the most robust recipe is absolute poses $+$ relative transport, no meta, and path-aware labels.

In the baseline comparison, under identical conditions (same candidate pool $\mathcal{G}$, endpoint–IK gate at $G(f)$, planner/timeouts, budget $B$, and execution template), Table~\ref{tab:baseline_runtime} reports executed-sim outcomes over 100 trials per method. The rank–and–select policy raises Succ@1 from $47\%$ ($47/100$) to $53\%$ ($53/100$). By two attempts, cumulative success increases from $62\%$ ($47{+}15$) to $83\%$ ($53{+}30$). Overall success within three attempts improves $74\%\!\to\!87\%$ (failure $26\%\!\to\!13\%$). Notably, successes concentrate in the first two calls (Succ@1/2 $=\ 83\%$ vs.\ $62\%$ for the baseline), indicating earlier wins and therefore better planner efficiency. The smaller Succ@3 count (4 vs.\ 12) reflects this front-loading rather than fewer total successes.

Most failures for the model predictions occur during the IK transition from $L_1$ to $L_2$. This is expected because the $L_1 \rightarrow L_2$ segment is introduced by the pairing step rather than the raw data collection, and the labels do not include an IK check on that segment.

\section{Conclusion and Future works}

This work investigates whether a dual-output MLP trained solely on scalable, physics-free endpoint-IK and mesh-sweep labels can guide executed trajectories. In executed simulation, the learned scores consistently improve Success@$K$ (earlier wins within the first two calls) and reduce expected planner calls per scene relative to a brute-force, rank-agnostic baseline. The current scope is a single rigid cuboid with side-face grasps and a fixed waypoint template. Labels are generated without physics, whereas execution includes physics and slip. A precomputed, face-balanced grasp pool is used and filtering is performed by the initial down face, excluding top/bottom grasps and clutter. Looking ahead, objects and scenes will be diversified, top/bottom grasps enabled where hardware permits, and training will incorporate execution-proximal signals (physics-aware sweeps, continuous path-collision checks, and logged planner outcomes).



\bibliographystyle{named}
\bibliography{references} 

\end{document}